\pdfoutput=1

\documentclass[11pt]{article}

\usepackage{EMNLP2022}

\usepackage{times}
\usepackage{latexsym}
\usepackage[T1]{fontenc}
\usepackage[utf8]{inputenc}
\usepackage{microtype}
\usepackage{inconsolata}
\usepackage{amsfonts}
\usepackage{amsmath}
\usepackage{amssymb}
\usepackage{xcolor}
\usepackage{blindtext}
\usepackage{graphicx}
\usepackage{booktabs}
\usepackage{multirow}
\usepackage{enumerate}
\usepackage{algorithm}
\usepackage[noend]{algpseudocode}
\usepackage{tabularx}
\usepackage{enumitem}
\usepackage{setspace}

\title{FaD-VLP: Fashion Vision-and-Language Pre-training \\ towards Unified Retrieval and Captioning}

\author{Suvir Mirchandani \\ Stanford University \\ \texttt{suvir@cs.stanford.edu}
        \And
        Licheng Yu \\ Meta AI \\ \texttt{lichengyu@meta.com}
        \And
        Mengjiao Wang \\ Meta AI \\ \texttt{mengjiaow@meta.com}
        \AND
        Animesh Sinha \\ Meta AI \\ \texttt{animeshsinha@meta.com}
        \And
        Wenwen Jiang \\ Meta AI \\ \texttt{wenwenj@meta.com}
        \And
        Tao Xiang \\ Meta AI / University of Surrey  \\ \texttt{txiang@meta.com}
        \AND
        Ning Zhang \\ Meta AI \\ \texttt{ningzhang@meta.com}}
        
\begin{document}
\newcommand{\placeholder}[1]{\textcolor{lightgray}{\blindtext[#1]}}
\newcommand{\sm}[1]{\textcolor{teal}{Suvir: #1}}

\newcommand{\E}{\mathbb{E}}
\newcommand{\B}{\mathbb{B}}
\newcommand{\argmax}{\text{argmax}}
\newcommand{\argmin}{\text{argmin}}
\renewcommand{\L}{\mathcal{L}}
\newcommand{\D}{\mathcal{D}}
\newcommand{\T}{\mathcal{T}}
\newcommand{\M}{\mathcal{M}}
\newcommand{\V}{\mathcal{V}}
\newcommand{\similarity}{\kappa}
\newcommand{\featI}{\phi_{I}}
\newcommand{\featT}{\phi_{T}}
\newcommand{\featM}{\phi_{M}}
\renewcommand{\i}{\mathbf{i}}
\renewcommand{\it}{\mathbf{i_t}}
\newcommand{\m}{\mathbf{m}}
\renewcommand{\t}{\mathbf{t}}
\newcommand{\Iref}{I_\textit{ref}}
\newcommand{\Tref}{T_\textit{ref}}
\newcommand{\Trel}{T_\textit{rel}}
\newcommand{\Itgt}{I_\textit{tgt}}
\newcommand{\Ttgt}{T_\textit{tgt}}

\newcolumntype{Y}{>{\centering\arraybackslash}X}

\maketitle
\begin{abstract}
\vspace{-0.02in}
\looseness=-1
\linespread{0.965}\selectfont
Multimodal tasks in the fashion domain have significant potential for e-commerce, but involve challenging vision-and-language learning problems---\textit{e.g.}, retrieving a fashion item given a reference image plus text feedback from a user.
Prior works on multimodal fashion tasks have either been limited by the data in individual benchmarks, or have leveraged generic vision-and-language pre-training but have not taken advantage of the characteristics of fashion data.
Additionally, these works have mainly been restricted to multimodal understanding tasks.
To address these gaps, we make two key contributions.
First, we propose a novel fashion-specific pre-training framework based on weakly-supervised triplets constructed from fashion image-text pairs.
We show the triplet-based tasks are an effective addition to standard multimodal pre-training tasks.
Second, we propose a flexible decoder-based model architecture capable of both fashion retrieval and captioning tasks. 
Together, our model design and pre-training approach are competitive on a diverse set of fashion tasks, including cross-modal retrieval, image retrieval with text feedback, image captioning, relative image captioning, and multimodal categorization.
\end{abstract}

\section{Introduction}
\label{sec:introduction}
Artificial intelligence has taken the fashion industry by storm in recent years. Significant advances have been made in tasks like recommendation \cite{mcauley2015image, deldjoo2022review} and virtual try-on \cite{han2018viton, yang2022full}. In addition to these primarily visual tasks, \textit{multimodal} tasks are of particular interest in fashion for e-commerce applications: for example, text-to-image retrieval enables a shopper to identify a desired clothing item via a language query \cite{zhuge2021kaleido}.

\begin{figure}
    \centering
    \includegraphics[width=\columnwidth]{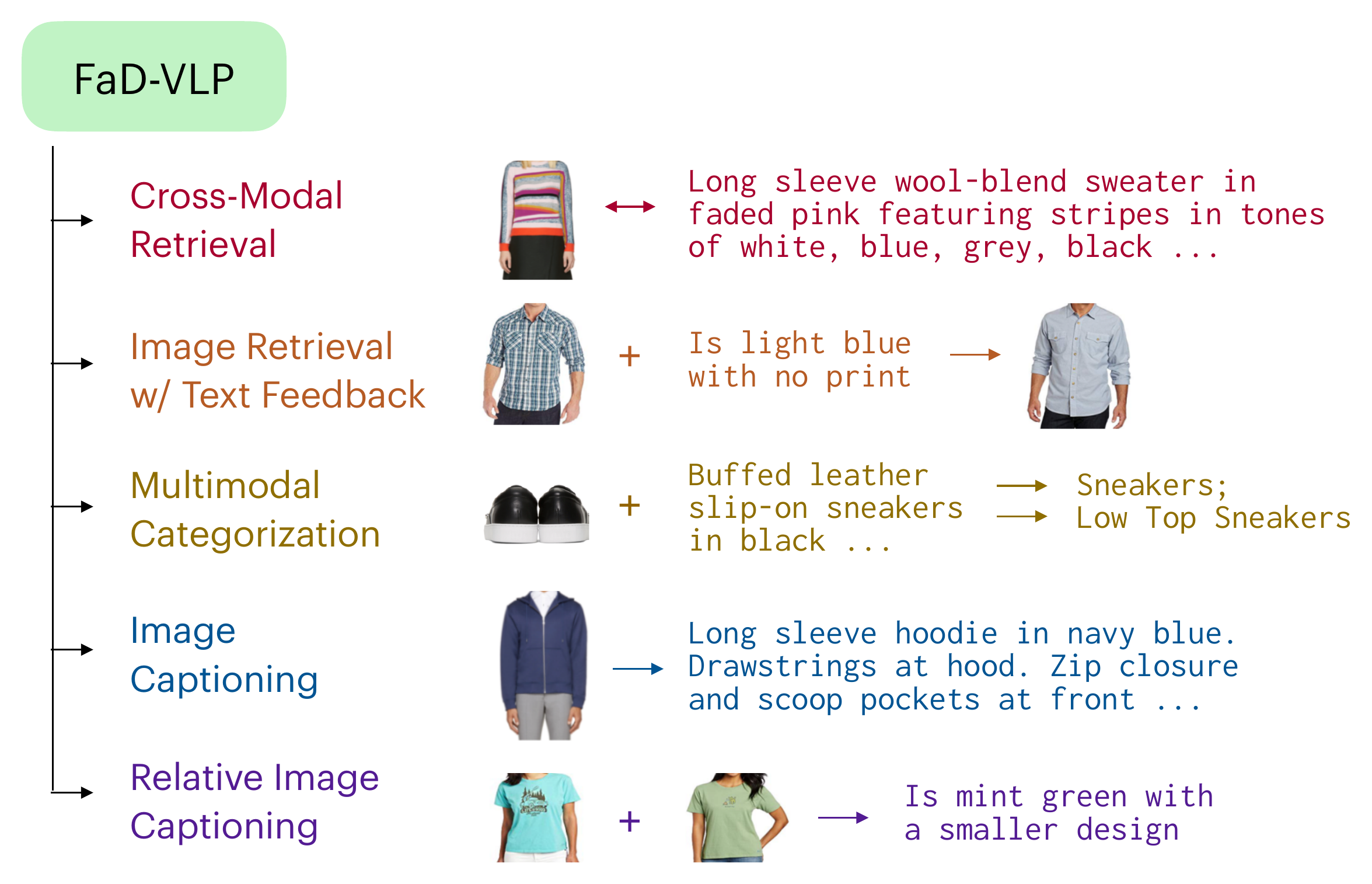}
    \caption{We present FaD-VLP, a flexible architecture and pre-training method that supports retrieval-based and captioning-based tasks in the fashion domain.}
    \label{fig:front-figure}

\end{figure}

A key opportunity to enhance customers' shopping experiences is in the development of \textit{interactive} multimodal shopping assistants, whereby a user could converse with a system to identify a desired product \cite{yuan2021conversational, han2022uigr}. 
As in Figure~\ref{fig:front-figure}, a smart assistant is desired to perform multiple diverse tasks, \textit{e.g.}, cross-modal retrieval, image retrieval with text feedback, multimodal categorization, image captioning, and relative image captioning.
Among them, perhaps the most notable task in fashion is image retrieval with text feedback, where the goal is to retrieve a target image given a reference image coupled with a user's language feedback (\textit{e.g.}, ``show me a similar shirt in light blue with no print'')~\cite{wu2021fashion, lee2021cosmo, kim2021dual}.
In addition to retrieval-based tasks, a central capability of conversational shopping assistants is in captioning-based tasks: describing items in detail \cite{yang2020fashion} or the differences among them. 
However, existing works on image retrieval with text feedback have almost exclusively studied that task in isolation, focusing on specialized architectures and fusion methods, with data limited by particular benchmarks \cite{lee2021cosmo, kim2021dual}.

To train a model that can perform well on several fashion-specific multimodal use cases, we observe an opportunity in the vast availability of multimodal fashion data on e-commerce platforms. 
While vision-language pre-trained (VLP) models have been highly successful for the general domain \cite{lu2019vilbert, li2020oscar, su2020vlbert}, prior work has suggested that general VLP models are helpful but suboptimal for the fashion domain~\cite{zhuge2021kaleido, liu2021image, goenka2022fashionvlp}. 
Fashion images represent a domain shift from the pre-training data \cite{liu2021image}, and fashion tasks often require fine-grained representations rather than coarse representations from general VLP models \cite{zhuge2021kaleido}.

To this end, we propose a domain-specific fashion pre-training procedure that takes advantage of fashion image-text data from multiple fashion catalogues. 
Our approach is inspired by the way that users might shop, via \textit{comparisons}: a user may first identify a product, express a desired change in language, and then look for a new product that better matches their preferences.
Given that data in this triplet form---reference product, modification, target product---is not nearly as common as the paired image-text data, we propose a lightweight method for constructing weakly-supervised pseudo-triplet data from image-text pairs. 
Additionally, we propose a unified, decoder-based model architecture for both retrieval-based and captioning-based fashion tasks. 
Together, we refer to our architecture and pre-training approach as FaD-VLP: Fashion Decoder with Vision-and-Language Pre-training.

To summarize, we make the following contributions. 
We propose a unified architecture for retrieval-based and captioning-based fashion tasks (Section \ref{sec:model-overview}) and a fashion pre-training framework, including 2 novel pre-training tasks based on weakly-supervised pseudo-triplets (Section \ref{sec:pre-training-objectives}). 
Our approach achieves competitive performance on 7 downstream fashion tasks: image-to-text retrieval, text-to-image retrieval, image retrieval with text feedback, category recognition, subcategory recognition, image captioning, and relative image captioning (Sections \ref{sec:experimental_setup} and \ref{sec:comparison-with-sota-models}). 
Finally, we conduct a thorough ablation study to analyze the effects of our pre-training procedure (Section \ref{sec:effect-of-pre-training-tasks}).

\section{Related Work}
\label{sec:related-work}
A substantial body of work has focused on using the Transformer architecture \cite{vaswani2017transformer} in the context of vision-and-language pre-training (VLP) \cite{li2019visualbert,su2020vlbert,chen2020uniter,radford2021clip,li2021align,yu2022coca}. 
Recent works have begun to focus on the fashion domain  \cite{gao2020fashionbert,zhuge2021kaleido,zhu2021k3m,dong2021m5product,zhang2021ufcbert,goenka2022fashionvlp,yu2022commercemm}. 
VLP works generally differ in their choice of model architecture and pre-training objectives.

\noindent \textbf{Model Architecture.}
Most existing VLP models, especially in the fashion domain, use encoder-style modules for both image and text, focusing on multimodal understanding tasks (which do not involve generation---\textit{e.g.}, image-text retrieval, multimodal classification).
There are two main classes of these models: ($i$) single-stream early fusion \cite{li2019visualbert,su2020vlbert,chen2020uniter,li2020oscar}, and ($ii$) two-stream late fusion \cite{tan2019lxmert,lu2019vilbert,jia2021align,radford2021clip}.
The nature of the downstream tasks often influences the choice of number of streams; \textit{e.g.}, image-text retrieval is most practical with late fusion architectures which can have faster inference.
In this work, we propose a flexible decoder-based model architecture, which embraces the advantage of both early and late fusion mechanisms, and is capable of not only multimodal understanding tasks, but also captioning tasks (\textit{e.g.}, image captioning and relative image captioning).

\begin{figure*}
    \centering
    \includegraphics[width=\textwidth]{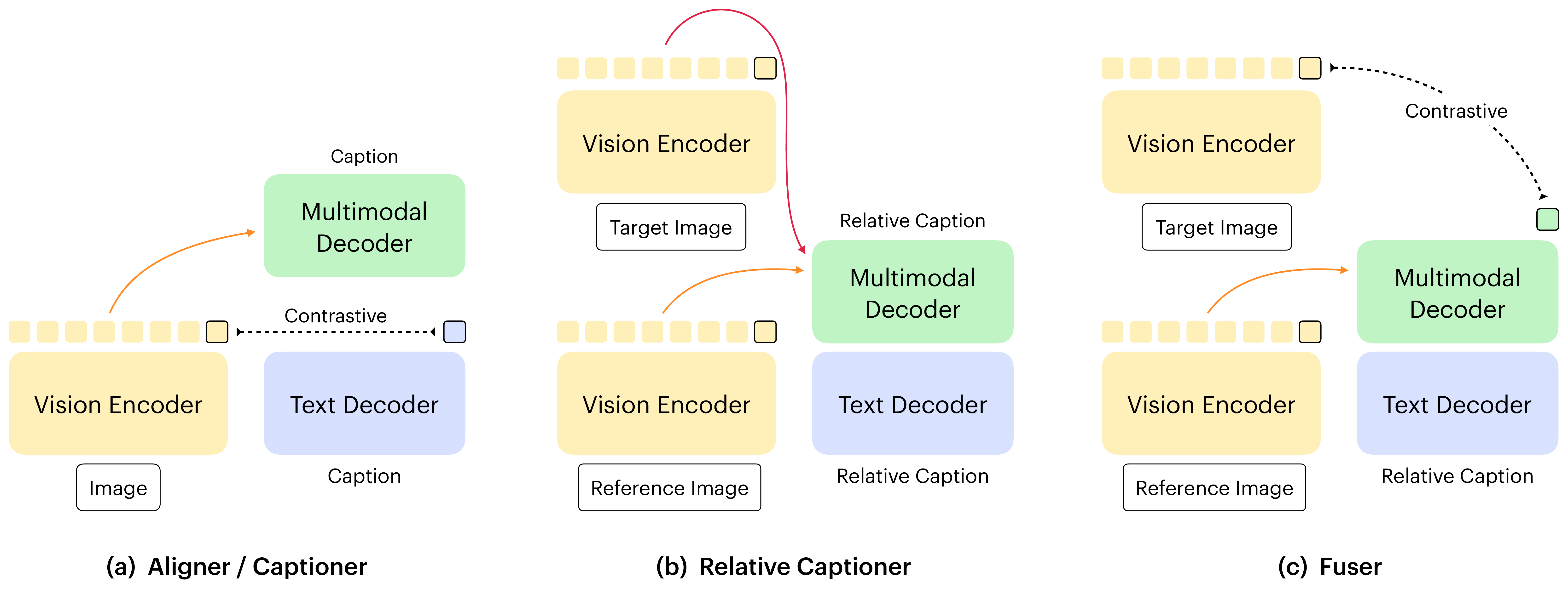}
\caption{Our proposed FaD-VLP architecture consists of an image encoder, a text decoder, and a multimodal decoder, with three configurations conforming to various retrieval and captioning tasks. Shared colors indicate shared parameters, curved arrows represent cross attention, and tokens with a bold border denote pooled representations.}
    \label{fig:architecture}
\end{figure*}

\noindent \textbf{Pre-training Objectives.} Several pre-training tasks have been effectively used for VLP.
Some of the most popular include masked modeling or matching objectives for the different modalities \cite{li2019visualbert,lu2019vilbert,su2020vlbert,chen2020uniter}; others include cross-modal contrastive learning \cite{li2021align,radford2021clip,li2021unimo}, caption generation \cite{zhou2020unifiedvlp,wang2021simvlm}, and object tagging \cite{li2020oscar}.
Fashion data has some unique properties that could be leveraged at pre-training, partly to mitigate the domain shift which makes generic VLP less effective for fashion \cite{zhuge2021kaleido}.
For example, there are more structured attributes in fashion captions, which entitles people to naturally do comparisons when choosing their desired shopping items.
Inspired by this, we propose that weak triplet-based comparison is used as the basis for additional pre-training tasks.

\section{Method}
\label{sec:method}
We introduce FaD-VLP, our architecture and pre-training method for fashion tasks. We first detail our architecture design (Figure~\ref{fig:architecture}), which unifies several retrieval and captioning settings. We then describe our pre-training approach. 

\subsection{Model Overview}
\label{sec:model-overview}
To motivate our model architecture, we enumerate three desired properties:

\begin{enumerate}[label=\roman*.,leftmargin=0cm,itemindent=.5cm,labelwidth=\itemindent,labelsep=0cm,align=left]
    \item \textit{Dual Image \& Text Encoders.} As referenced in Section \ref{sec:related-work}, two-stream / dual-encoder architectures are more efficient for cross-modal retrieval than single-stream architectures. With dual encoders, candidate embeddings can be retrieved using a lightweight similarity function (\textit{e.g.}, dot product) with a particular query embedding.
    \item \textit{Dual Multimodal \& Text Encoders.} Key to our pre-training procedure is the alignment of multimodal representations with image representations. This setup is useful for the downstream task of image retrieval with text feedback: a target image is retrieved given an image with text feedback. We desire an architecture that is dual-stream with respect to a hybrid-modal input (image and text) and another image.
    \item \textit{Multimodal Decoder for Text Generation.} For captioning tasks, we need to generate text given image input. Thus, we desire that the architecture contains a multimodal decoder.
\end{enumerate}

To satisfy (i) and (iii), prior work \cite{li2022blip} has used a mixture of unimodal encoders and encoder-decoders; more recently, \citet{yu2022coca} demonstrated the effectiveness of using single decoder-based model; a decoder can be used for generation, but can also provide global representations given a whole sequence.

Building upon this result, our architecture is decoder-based, and consists of three  modules: a visual encoder $\V$, a text decoder $\T$, and a multimodal decoder $\M$. For $\V$, we use a convolutional network. We obtain image token representations from the intermediate outputs of the convolutional network (\textit{i.e.}, the output of layers 3 and 4 in a ResNet-50, following \citet{kim2021dual}). We obtain pooled representations from $\V$ using average pooling over the final feature map. We use a multi-layer transformer architecture for $\T$ and $\M$. Each layer consists of a causal multi-headed self-attention module followed by a feed-forward network and layer normalization. For $\M$, we also include a cross-attention layer between the image representation and the outputs of the causal self-attention.  We extract pooled representations from $\T$ or $\M$ using the output of corresponding to an $\texttt{[EOS]}$ token (which has attended to all prior tokens).

Our architecture has the following modes:
\begin{enumerate}[label=(\alph*),leftmargin=0cm,itemindent=.6cm,labelwidth=\itemindent,labelsep=0cm,align=left]
    \item \textbf{Aligner / Captioner.} This mode can align cross-modal representations or caption an image. For alignment, we input a caption to $\T$ and an image to $\V$, extracting the pooled representations. For captioning, we pass the outputs of $\T$ to $\M$ and condition $\M$ on the image  via cross attention.
    \item \textbf{Relative Captioner.} In this mode, we can input a text (\textit{e.g.}, a relative caption comparing two images) into $\T$ and two image representations into $\M$. We use a gated, second cross-attention mechanism for the second image input. The second cross attention module is computed on the output of the first cross attention. The multimodal decoder can be trained to generate a relative caption conditioned on images.
    \item \textbf{Fuser.} This mode can fuse an image and text, and align the results with another image. We input a relative caption to $\T$, a reference image representation to $\M$ via cross attention, and a target image into $\V$. The pooled representation of the target can be aligned with the fused representation from $\M$.
\end{enumerate}

The three modes repurpose the same components which allows us to share parameters among modes. We prepend text inputs with a special token indicating which of the three modes the architecture is operating in.
Following prior VLP work \cite{lu2019vilbert, li2020oscar, su2020vlbert, li2021align, li2022blip}, we initialize our model with BERT encoder weights \cite{devlin2019bert}. As BERT does not have cross-attention parameters, they are learned from scratch. 

\subsection{Pre-training Objectives}
\label{sec:pre-training-objectives}
Image-text pairs are a common choice for pre-training vision-language systems; these pairs can be mined or repurposed from existing datasets, such as captioning datasets. As described in Section \ref{sec:datasets}, we repurpose a set of fashion datasets to form a pre-training dataset $\D$ consisting of fashion image-text pairs. We use $\D$ for domain-specific pre-training. This section describes our pre-training tasks, which include two tasks based on paired data (Section \ref{sec:pre-training-with-pairs}), as well as two novel tasks based on triplet data (Section \ref{sec:pre-training-with-triplets}). Implementation details are given in Appendix~\ref{app:training-details}.

\subsubsection{Pre-training with Pairs}
\label{sec:pre-training-with-pairs}
\paragraph{Cross-Modal Contrastive Learning (CMC).}

To align the representations of images with their corresponding texts, we use a cross-modal contrastive loss. Given an image $I$ and a text $T$, we extract the pooled feature vectors $\i$ and $\t$ from the visual encoder $\V(I)$ and the text encoder $\T(T)$. We then project $\i$ and $\t$ to a normalized lower-dimensional joint embedding space using two learned linear transformations, $f$ and $g$. We measure the similarity between $\i$ and $\t$ as
\begin{equation*}
    \similarity(\i, \t) = f(\i)^T g(\t).
\end{equation*}
We can push the embeddings of matched images and texts together according to this similarity metric, and unmatched embeddings further apart, by applying the following bidirectional InfoNCE loss \cite{oord2018representation, zhang2020contrastive}:
\begin{align*}
    \L_{\text{CMC}}= &-\frac{1}{B}\sum_{j=1}^{B} \Big(
                     \log \frac{\exp(\similarity(\i^{(j)}, \t^{(j)}))}{\sum_{k=1}^B \exp(\similarity(\i^{(j)}, \t^{(k)}))} \\
                   + &\log \frac{\exp(\similarity(\i^{(j)}, \t^{(j)}))}{\sum_{k=1}^B \exp(\similarity(\i^{(k)}), \t^{(j)}))} \Big).
\end{align*}
where $B$ is a sample of indices from our pretraining dataset $\D$.
We use the Aligner / Captioner mode of our architecture for this task.
\paragraph{Image Caption Language Modeling (ICLM).}

In addition to alignment, we encourage the model to gain image-grounded text generation capabilities. We use a language modeling loss that maximizes the conditional probability of a caption $T^{(j)}$ given an image $I^{(j)}$:
\begin{equation*}
    \L_{\text{ICLM}} = -\frac{1}{B} \sum_{j=1}^{B} \sum_{k=1}^{|T^{(j)}|} \log P_{\M}(T^{(j)}_k \mid T^{(j)}_{<k}, I^{(j)})
\end{equation*}
where $T^{(j)}_k$ refers to the $k$th token, and $T^{(j)}_{<k}$ refers to the context history of tokens.
As with the CMC task, we use the Aligner / Captioner mode of our architecture.

\subsubsection{Pre-training with Triplets}
\label{sec:pre-training-with-triplets}
Taking inspiration from the way users utilize \textit{comparisons} when browsing products (e.g., looking at a product, having a desired change in mind, and identifying a new product), we hypothesize that (image, text, image) \textit{triplets} can be used to build additional multimodal capabilities into the model beyond cross-modal tasks like CMC and ICLM.

Below, we describe two pre-training objectives that assume access to triplets $(\Iref, \Trel, \Itgt)$, where $\Itgt$ is a target image and $\Trel$ is a relative caption describing the difference between $\Itgt$ and some reference image $\Iref$. Note we only have access to image-text pairs in $\D$; we describe how we construct these triplets from pairs in Section \ref{sec:constructing-pseudo-triplets}.

\paragraph{Hybrid-Modal Contrastive Learning (HMC).}

We propose a pre-training task, using the Fuser mode of the architecture, that aligns the fused representation of $\Iref$ and $\Trel$ with the unimodal representation of $\Itgt$. Intuitively, this would imbue the model with the ability to modify $\Iref$ in embedding space, as specified by $\Trel$.

We project the pooled features $\m$ and $\it$ of the multimodal and target image embeddings $\M(\Iref, \Trel)$ and $\V(\Itgt)$ using two learned linear transformations, $h$ and $f$ respectively. We measure the similarity between $\m$ and $\it$ as
\begin{equation*}
    \similarity'(\m, \it) = h(\m)^T f(\it).
\end{equation*}
We then apply the following contrastive loss: 
\begin{align*}
    \L_{\text{HMC}}= &-\frac{1}{B}\sum_{j=1}^{B} 
                     \log \frac{\exp(\similarity'(\m^{(j)}, \it^{(j)}))}{\sum_{k=1}^B \exp(\similarity'(\m^{(j)}, \it^{(k)}))}.
\end{align*}

\paragraph{Relative Caption Language Modeling (RCLM).}

We additionally apply a language modeling loss, using the Relative Captioner mode of our architecture, that maximizes the conditional probability of a relative caption given a reference and a target image:
\begin{align*}
    \L_{\text{RCLM}} =& -\frac{1}{B} \sum_{j=1}^{B} \sum_{k=1}^{|\Trel^{(j)}|} \\
    &\log P_{\M}(T^{(j)}_{\textit{rel},k} \mid T^{(j)}_{\textit{rel},<k}, \Iref, \Itgt).
\end{align*}

\subsection{Constructing Pseudo-Triplets}
\label{sec:constructing-pseudo-triplets}
\begin{figure}
    \centering
    \includegraphics[width=\columnwidth]{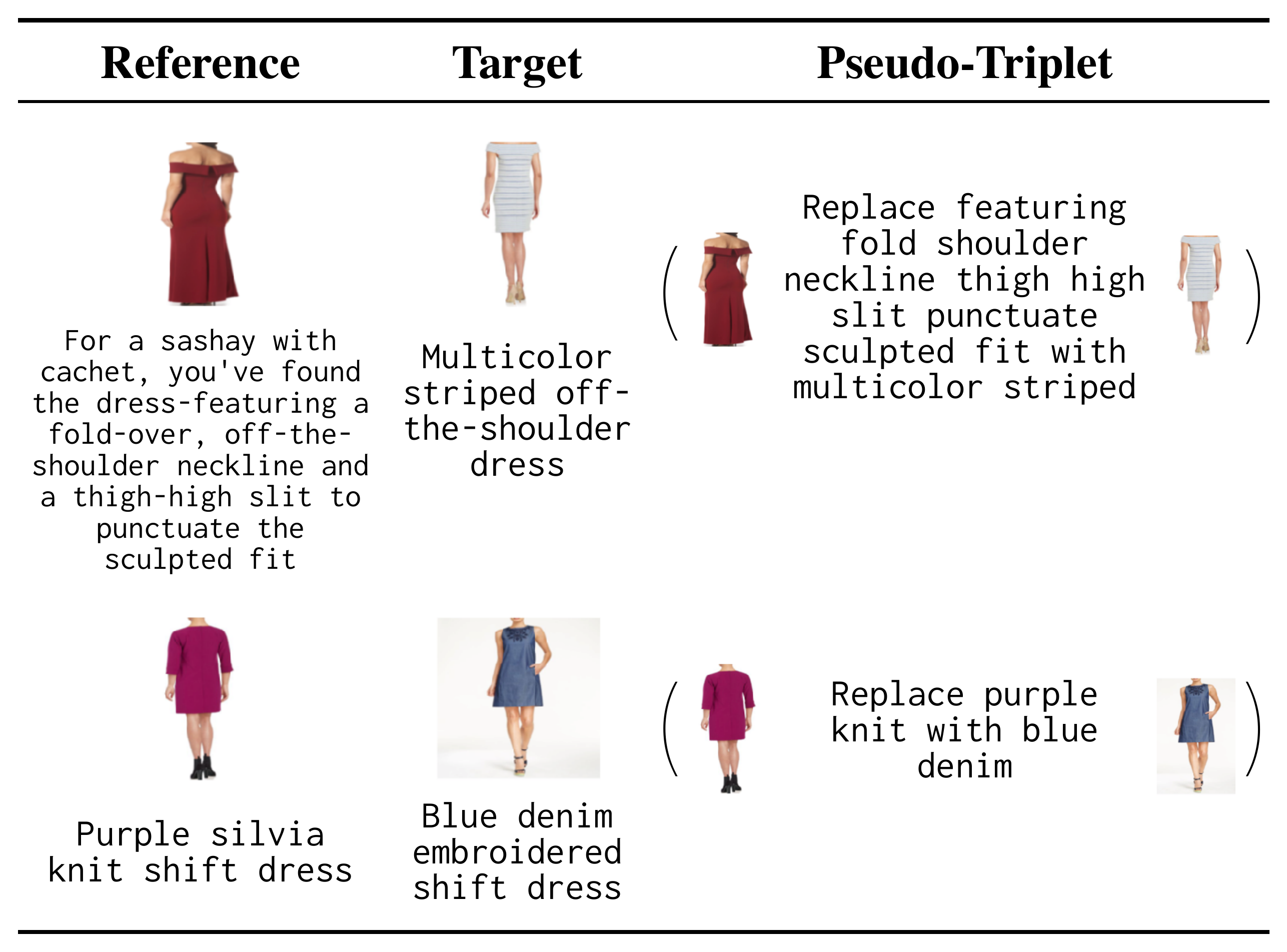}
    \caption{Two examples of pseudo-triplets. From a reference image-text pair, we find a target image-text pair and then construct a relative caption for the two images.}
    \label{fig:pseudo-triplets}
\end{figure}

Collecting additional pre-training data for the triplet-based tasks would be expensive, thus we aim to construct triplets purely from the paired image-text data. We propose a simple approach for generating weakly supervised pseudo-triplets.

We iterate through each image-text pair in $\D$, treating it as a reference $(\Iref, \Tref)$. 
As in Figure~\ref{fig:pseudo-triplets}, for each reference, we select a target image-text pair $(\Itgt, \Ttgt)$, then construct a relative caption as a function $rel$ of the two captions, $\Tref$ and $\Ttgt$. We use $(\Iref, rel(\Tref, \Ttgt), \Itgt)$ as our pseudo-triplet.

\paragraph{Selecting Targets.}
Intuitively, we want $\Iref$ and $\Itgt$ to be related (such that there are some shared properties of the reference and the target item) but not identical (so the relative caption would be meaningful). To do this, for a given reference $(\Iref,\Tref)$, we find
\begin{equation*}
    \argmin_{j \in S} { \Delta ( (\Iref,\Tref), (I^{(j)},T^{(j)}) ) }
\end{equation*}
\noindent
where $\Delta$ is a similarity metric and $S$ is a set of indices sampled from the indices in $\D$, and set $(\Itgt, \Ttgt) = (I^{(j)}, T^{(j)})$. We use the following metric:
\begin{align*}
    \Delta( (I, T), (I', T') ) = &-\lambda_1 \cdot \similarity( \featI(I), \featI(I'))) \\
                                 &-  \lambda_2 \cdot \similarity(\featI(T), \featT(T')) \\
                                 &+  \lambda_3 \cdot d(T, T'),
\end{align*}
where $\featI$ and $\featT$ are feature extractors for image and text, $d$ is the token-wise Hamming distance (for nouns, adjectives, and participles), $\kappa$ is cosine similarity, and $\lambda_1$, $\lambda_2$, and $\lambda_3$ are scalar weights. 
We use frozen feature extractors: ResNet-50 \cite{he2016deep} for $\featI$ and all-MiniLM-L12-v2 \cite{reimers2019sentence} for $\featT$. The intuition behind our choice for $\Delta$ is that pairings should be visually similar, and texts should be semantically similar and have token overlap.

\begin{table}[]
    \centering
    \small
    \begin{tabular}{lc}
\toprule
                            \textbf{Name} & \textbf{\# Pairs} \\
\midrule
             FACAD \cite{yang2020fashion} &              888K \\
Fashion-Gen \cite{rostamzadeh2018fashion} &              260K \\
      Fashion200K \cite{han2017automatic} &              172K \\
      Shopping100K \cite{ak2018efficient} &              100K \\
           DeepFashion \cite{liu2016deep} &               26K \\
\bottomrule
\end{tabular}
    \caption{Breakdown of the pre-training dataset.}
    \label{tab:datasets}
\end{table}

\begin{table*}[]
    \centering
    \small
\begin{tabularx}{\textwidth}{p{125pt}YYYYYYY}
\toprule
     \multirow{2}{*}{\textbf{Method}} & \multicolumn{3}{c}{\textbf{Image-to-Text}} & \multicolumn{3}{c}{\textbf{Text-to-Image}} & \multirow{2}{*}{\textbf{Average}} \\
                                      &           \textbf{R@1} &    \textbf{R@5} &   \textbf{R@10} &           \textbf{R@1} &    \textbf{R@5} &   \textbf{R@10} &                                   \\
\midrule
         VSE \cite{kiros2014unifying} &                   4.01 &           11.03 &           22.14 &                   4.35 &           12.76 &           20.91 &                             12.53 \\
         VSE++ \cite{faghri2018vsepp} &                   4.59 &           14.99 &           24.10 &                   4.60 &           16.89 &           28.99 &                             15.69 \\
           SCAN \cite{lee2018stacked} &                   4.59 &           16.50 &           26.60 &                   4.30 &           13.00 &           22.30 &                             14.55 \\
         PFAN \cite{wang2019position} &                   4.29 &           14.90 &           24.20 &                   6.20 &           20.79 &           31.52 &                             16.98 \\
         CLIP, 0-shot \cite{radford2021clip} & 18.56 & 37.86 & 51.02 & 16.78 & 37.92 & 50.22 & 35.39\\
         ViLBERT \cite{lu2019vilbert} &                  20.97 &           40.49 &           48.21 &                  21.12 &           37.23 &           50.11 &                             36.35 \\
           VLBERT \cite{su2020vlbert} &                  19.26 &           39.90 &           46.05 &                  22.63 &           36.48 &           48.52 &                             35.47 \\
     ImageBERT \cite{qi2020imagebert} &                  22.76 &           41.89 &           50.77 &                  24.78 &           45.20 &           55.90 &                             40.22 \\
FashionBERT \cite{gao2020fashionbert} &                  23.96 &           46.31 &           52.12 &                  26.75 &           46.48 &           55.74 &                             41.89 \\
             OSCAR \cite{li2020oscar} &                  23.39 &           44.67 &           52.55 &                  25.10 &           49.14 &           56.68 &                             41.92 \\
 Kaleido-BERT \cite{zhuge2021kaleido} &                  27.99 &           60.09 &           68.37 &                  33.88 &           60.60 &           68.59 &                             53.25 \\
FaD-VLP (Ours, \textit{w/o} CLIP init.) &                  59.88 &           83.64 &           91.52 &                  55.24 &           83.18 &           91.30 &                             77.46 \\
                       FaD-VLP (Ours) &         \bfseries 64.30 & \bfseries 86.78 & \bfseries 93.48 &        \bfseries 58.66 & \bfseries 84.92 & \bfseries 91.58 &                   \bfseries 79.95 \\
\bottomrule
\end{tabularx}
    \caption{Results for Image-Text / Text-Image Retrieval (ITR / TIR) on Fashion-Gen \cite{rostamzadeh2018fashion}.}
    \label{tab:fashiongen-retrieval}
\end{table*}

\begin{table*}[]
    \centering
    \small
\begin{tabularx}{\textwidth}{p{145pt}YYYYYYY}
\toprule
       \multirow{2}{*}{\textbf{Method}} & \multicolumn{2}{c}{\textbf{Dress}} & \multicolumn{2}{c}{\textbf{Shirt}} & \multicolumn{2}{c}{\textbf{Toptee}} & \multirow{2}{*}{\textbf{Average}} \\
                                        &   \textbf{R@10} &   \textbf{R@50} &   \textbf{R@10} &   \textbf{R@50} &   \textbf{R@10} &   \textbf{R@50} &                                   \\
\midrule
            TIRG \cite{vo2019composing} &           14.13 &           34.61 &           13.10 &           30.91 &           14.79 &           34.37 &                             23.66 \\
           CIRPLANT \cite{liu2021image} &           17.45 &           40.41 &           17.53 &           38.81 &           21.64 &           45.38 &                             30.20 \\
              CoSMo \cite{lee2021cosmo} &           21.39 &           44.45 &           16.90 &           37.49 &           21.32 &           46.02 &                             31.25 \\
 FashionVLP \cite{goenka2022fashionvlp} &           26.77 &           53.20 &           22.67 &           46.22 &           28.51 &           57.47 &                             39.14 \\
               DCNet \cite{kim2021dual} &           28.95 &           56.07 &           23.95 &           47.30 &           30.44 &           58.29 &                             40.84 \\
FaD-VLP (Ours, \textit{w/o} CLIP init.) &           29.15 &           55.97 &           23.45 &           46.61 &           30.85 &           57.57 &                             40.60 \\
                         FaD-VLP (Ours) & 32.08 & 57.96 & 25.22 & 49.71 &  33.20 & 60.84 & 43.17 \\
             Prog.~Lrn.~-~RN-50 \cite{zhao2022progressive}$^\dagger$ & {29.00} & {53.94} & {35.43} & {58.88} & {39.16} & {64.56} & {46.83} \\
            Prog.~Lrn.~-~ViT-B/32 \cite{zhao2022progressive}$^\dagger$ & \textbf{33.60} & \textbf{58.90} & \textbf{39.45} & \textbf{61.78} & \textbf{43.96} & \textbf{68.33} & \textbf{51.01} \\
\bottomrule
\end{tabularx}

    \caption{Results for Image Retrieval with Text Feedback (IRTF) on Fashion IQ \cite{wu2021fashion}. $^\dagger$ refers to concurrent work on fashion vision-language pre-training.}
    \label{tab:fashioniq-retrieval}
    \vspace{-5pt}
\end{table*}

\paragraph{Constructing Relative Captions.}
\label{ref:constructing-relative-captions}

To construct a relative caption $\Trel$, we design a function $rel$ of the reference and target captions $\Tref$ and $\Ttgt$. Our goal is to describe the difference between both images such that HMC and RCLM are empirically helpful for our downstream tasks. By training $\M$ to match the fused representation of an image $\Iref$ and relative caption $\Trel$, we would be adding  fusion capability to $\M$ (with no extra data) and potentially getting stronger representations of image and text.

We use a simple and lightweight procedure for the \textit{rel} function. We extract the first sentence from $\Tref$ and $\Ttgt$, and perform part-of-speech tagging. We filter out all tokens that do not function as nouns and adjectives in the caption, as well as all tokens that occur less than 500 times over the entire dataset. This leaves us with a noisy list of ``attributes'' for each image.

Next, we remove tokens that overlap between reference tokens and target tokens. This step removes some of the redundancy (at the token level) between the reference and target captions, as relative captions would not need to mention aspects in the target image that are unchanged from the reference image.

We use the remaining reference tokens and target tokens to fill a randomly selected template of the form ``\texttt{change <ref\_tokens> to <tgt\_tokens>},'' ``\texttt{<tgt\_tokens> instead of <ref\_tokens>},'' etc.
While our goal is not to produce fully grammatical or complete relative captions, we found most constructed sentences are meaningful, as exemplified in Figure~\ref{fig:pseudo-triplets}. More importantly, they are effective at improving pre-training.

\subsubsection{Bootstrapping with Relative Caption Generations}
The RCLM pre-training task enables the model to produce relative captions given two images. This gives us the ability to sample new relative captions at training time as a form of data augmentation. 
We use nucleus sampling (with $p=0.9$) to generate more diverse relative captions. 
Functionally, this step connects the Relative Captioner mode and the Fuser mode of our architecture.

\section{Experiments}
\label{sec:experimental_setup}
In this section, we provide the details of our pre-training dataset and downstream fashion tasks. Implementation details are given in Appendix \ref{app:training-details}.

\subsection{Datasets}
\label{sec:datasets}
Our pre-training dataset is comprised of image-text pairs from five fashion datasets: FACAD \cite{yang2020fashion}, Fashion-Gen \cite{rostamzadeh2018fashion}, Fashion200K \cite{han2017automatic},  Shopping100k \cite{ak2018efficient}, and DeepFashion \cite{liu2016deep}. Each of these datasets contains data sourced from fashion catalogues.  In total, our pre-training dataset consists of 1.4M image-text pairs, with the breakdown as listed in Table \ref{tab:datasets}. Further description is given in Appendix \ref{app:training-details}.

Following \citet{gao2020fashionbert} and \citet{zhuge2021kaleido}, we use the Fashion-Gen dataset for our cross-modal retrieval, captioning, and multimodal categorization tasks. We use Fashion IQ \cite{wu2021fashion} for our image retrieval with text feedback task and our relative captioning task. Fashion IQ contains 18K (reference image, text feedback, target image) training triplets and 6016 validation triplets over three categories: Dress, Shirt, and Toptee. Each (reference image, target image) pair is human-annotated with two relative captions, which are concatenated together \cite{wu2021fashion}.

\begin{table}[]
    \centering
    \small
    \begin{tabularx}{\columnwidth}{p{69pt}YYYYYY}
\toprule
                 \multirow{2}{*}{\textbf{Method}} & \multicolumn{3}{c}{\textbf{Category}} & \multicolumn{3}{c}{\textbf{Subcategory}} \\
                                                  &     \textbf{Acc.} &    \textbf{F1} &  \textbf{Avg.} &        \textbf{Acc.} &    \textbf{F1} &  \textbf{Avg.} \\
\midrule
                 ImageBERT \newline \cite{qi2020imagebert} &             90.77 &           69.9 &           80.3 &                80.11 &           57.5 &           75.0 \\
            FashionBERT \newline \cite{gao2020fashionbert} &             91.25 &           70.5 &           80.9 &                85.27 &           62.0 &           77.9 \\
                         OSCAR \newline \cite{li2020oscar} &             91.79 &           72.7 &           82.2 &                84.23 &           59.1 &           78.5 \\
             Kaleido-BERT \newline \cite{zhuge2021kaleido} &             95.07 &           71.4 &           83.2 &                88.07 &           63.6 &           79.7 \\
FaD-VLP (Ours) \newline [\textit{w/o} CLIP init.] &             97.90 &           89.3 &           93.6 &      \bfseries 93.53 & \bfseries 83.2 &           91.4 \\
                                   FaD-VLP (Ours) &   \bfseries 98.32 & \bfseries 89.5 & \bfseries 93.9 &                93.37 & \bfseries 83.2 & \bfseries 91.5 \\
\bottomrule
\end{tabularx}
    \caption{Results for Category / Subcategory Recognition (CR / SR) on Fashion-Gen \cite{wu2021fashion}.}
    \label{tab:fashiongen-categorization}
\end{table}

\subsection{Downstream Tasks}
We fine-tune the pre-trained model on 7 downstream tasks, which are defined as follows.

\noindent \textbf{Image-to-Text Retrieval (ITR) and Text-to-Image Retrieval (TIR).} Given a gallery $\Theta_{I,T}$ of fashion image-text pairs $(I,T)$, and a query image $I_q$ (or query text $T_q$), retrieve the corresponding text $T_q$ (or image $I_q$). We train these tasks with the CMC loss.

\noindent \textbf{Image Retrieval with Text Feedback (IRTF)}. Given a gallery $\Theta_I$ of fashion images $I$, a reference image $\Iref$, and a relative caption $\Trel$, retrieve the target image $\Itgt$ that most closely applies $\Trel$ to $\Iref$. We train this task with the HMC loss.

\noindent \textbf{Category Recognition (CR) and Subcategory Recognition (SR).} Given an image $I_q$ and a list of categories (or subcategories) $C$, predict the category (or subcategory) $c$ into which $I_q$ falls. Example categories include \{\texttt{SNEAKERS, JEANS}\} and and example subcategories include \{\texttt{SILKS} \texttt{\&} \texttt{CASHMERES}, \texttt{HEELED SANDALS}\}. We train these tasks with categorical cross entropy loss.

\noindent \textbf{Image Captioning (IC).} Given a fashion image $I_q$, generate a descriptive caption $T_q$. We train with the ICLM loss.

\noindent \textbf{Relative Image Captioning (RIC).} Given a reference image $\Iref$ and a target image $\Itgt$, generate a text $\Trel$ describing $\Itgt$ relative to $\Iref$. We train this task with the RCLM loss.

\paragraph{Evaluation Metrics}

For ITR, TIR, and IRTF, we follow prior work \cite{kim2021dual, zhuge2021kaleido} and evaluate the retrieval performance using Recall@$K$, or the percentage of queries for which the correct target is retrieved within the top $K$ results. For CR and SR, we follow \cite{zhuge2021kaleido} and report accuracy and the macro-F1 score over all categories (or subcategories). For IC and RIC, we report BLEU-4 \cite{papineni2002bleu}, METEOR \cite{banerjee2005meteor}, ROUGE-L \cite{lin2004rouge}, and CIDEr \cite{vedantam205cider}, abbreviated B, M, R, and C respectively; C is rescaled from 0--10 to to 0--100 before calculating the aggregate Sum metric. In our ablation study (Table \ref{tab:ablation-pre-training}), we use the aggregate metrics for each task.

\section{Results \& Discussion}
\label{sec:results_discussion}
\begin{table}[]
    \centering
    \small
    \begin{tabularx}{\columnwidth}{p{70pt}YYYYY}
\toprule
                             \textbf{Method} &     \textbf{B} &     \textbf{M} &     \textbf{R} &     \textbf{C} &    \textbf{Sum} \\
\midrule
       FashionBERT \newline \cite{gao2020fashionbert} &            3.3 &            9.8 &           29.7 &            0.3 &            45.8 \\
                    OSCAR \newline \cite{li2020oscar} &            4.5 &           10.9 &           30.1 &           0.31 &            48.6 \\
        Kaleido-BERT \newline \cite{zhuge2021kaleido} &            5.7 &           12.8 &           32.9 &           0.33 &            54.7 \\
FaD-VLP (Ours) \newline [\textit{w/o} CLIP init., PT] &           29.3 &           28.6 &           54.5 &           1.41 &           126.6 \\
            FaD-VLP (Ours) \newline [\textit{w/o} PT] &           31.0 &           29.6 &           55.8 &           1.55 &           131.9 \\
                              FaD-VLP (Ours) & \bfseries 31.1 & \bfseries 29.7 & \bfseries 56.0 & \bfseries 1.56 & \bfseries 132.4 \\
\bottomrule
\end{tabularx}

    \caption{Results for Image Captioning (IC) on Fashion-Gen \cite{rostamzadeh2018fashion}.}
    \label{tab:fashiongen-captioning}
\end{table}

\begin{table}[]
    \centering
    \small
    \begin{tabularx}{\columnwidth}{p{82pt}YYYYY}
\toprule
                            \textbf{Method} &     \textbf{B} &     \textbf{M} &     \textbf{R} &    \textbf{C} &   \textbf{Sum} \\
\midrule
                           Decoder Baseline &           13.2 &           17.3 &           38.9 &           0.7 &           76.5 \\
\textit{w/o} CLIP init, Triplet PT, Pair PT &           13.3 &           17.4 &           39.0 &          0.71 &           76.7 \\
           \textit{w/o} Triplet PT, Pair PT &           13.4 &           17.6 &           39.2 &          0.73 &           77.5 \\
                    \textit{w/o} Triplet PT &           13.8 &           18.4 &           39.6 &          0.78 &           79.5 \\
                             FaD-VLP (Ours) & \bfseries 14.5 & \bfseries 18.6 & \bfseries 40.8 & \bfseries 0.8 & \bfseries 82.0 \\
\bottomrule
\end{tabularx}
    \caption{Results for Relative Image Captioning (RIC) on Fashion IQ \cite{wu2021fashion}.}
    \label{tab:fashioniq-relative-captioning}
\end{table}

\begin{table*}[]
    \centering
    \small
    \begin{tabularx}{\textwidth}{p{10pt}ccccccYYYYYYYYY}
\toprule
\textbf{} & \textbf{CLIP} & \textbf{CMC} & \textbf{ICLM} & \textbf{HMC} & \textbf{RCLM} & \textbf{Boot} &  \textbf{Sum} &  \textbf{IRTF} &  \textbf{ITR} &  \textbf{TIR} &  \textbf{CR} &  \textbf{SR} &  \textbf{IC} &  \textbf{RIC} \\
\midrule
      (1) &               &              &               &              &               &               &        528.73 &          31.93 &         54.97 &         56.97 &        93.67 &        87.86 &       126.59 &         76.74 \\
      (2) &               &            \checkmark &             \checkmark &              &               &               &        546.09 &          38.00 &         58.18 &         61.14 &        94.50 &        89.38 &       125.52 &         79.37 \\
      (3) &               &            \checkmark &             \checkmark &            \checkmark &               &               &        546.48 &          39.83 &         58.38 &         61.48 &        93.03 &        88.73 &       125.39 &         79.64 \\
      (4) &               &            \checkmark &             \checkmark &            \checkmark &             \checkmark &               &        551.51 &          39.96 &         59.26 &         62.04 &        93.95 &        88.73 &       125.70 &         81.87 \\
      (5) &               &            \checkmark &             \checkmark &            \checkmark &             \checkmark &             \checkmark &        552.07 &          40.60 &         58.82 &         61.83 &        93.58 &        89.45 &       126.07 &         81.72 \\
      (6) &             \checkmark &              &               &              &               &               &        546.62 &          35.62 &         58.55 &         60.58 &        93.74 &        88.77 &       131.87 &         77.49 \\
      (7) &             \checkmark &            \checkmark &               &              &               &               &        551.86 &          37.20 &         59.68 &         63.00 &        93.20 &        88.20 &       131.97 &         78.61 \\
      (8) &             \checkmark &            \checkmark &             \checkmark &              &               &               &        562.12 &          40.66 &         61.18 &         64.42 &        94.25 &        89.29 &       132.78 &         79.54 \\
      (9) &             \checkmark &            \checkmark &             \checkmark &            \checkmark &               &               &        564.38 &          42.20 &         62.34 &         66.37 &        93.83 &        88.64 &       132.05 &         78.95 \\
     (10) &             \checkmark &            \checkmark &             \checkmark &            \checkmark &             \checkmark &               &        566.17 &          42.52 &         61.03 &         65.48 &        93.93 &        88.55 &       132.54 &         82.12 \\
     (11) &             \checkmark &            \checkmark &             \checkmark &            \checkmark &             \checkmark &             \checkmark &        569.81 &          43.17 &         62.87 &         67.05 &        94.05 &        88.30 &       132.39 &         81.98 \\
\bottomrule
\end{tabularx}
    \caption{Ablation on our pre-training objectives, with the aggregation metrics for each of our 7 downstream tasks and a meta-sum. CLIP refers to CLIP initialization; CMC, ICLM, HMC, and RCLM are pre-training tasks; Boot refers to bootstrapping.}
    \label{tab:ablation-pre-training}
\end{table*}

This section describes our results on our downstream tasks and an ablation study.

\subsection{Comparison with SOTA Models}
\label{sec:comparison-with-sota-models}
We compare FaD-VLP to existing work on our set of 7 downstream tasks: ITR/TIR (Table \ref{tab:fashiongen-retrieval}), IRTF (Table \ref{tab:fashioniq-retrieval}), CR/SR (Table \ref{tab:fashiongen-categorization}), IC (Table \ref{tab:fashiongen-captioning}), and RIC (Table \ref{tab:fashioniq-relative-captioning}). For ITR and TIR, FaD-VLP outperforms prior methods by a large margin, including methods with generic VLP \cite{li2020oscar}. We also include zero-shot retrieval results with a CLIP ResNet-50 encoder as a baseline. We show results of FaD-VLP with and without CLIP visual encoder initialization, indicating that the gain from fashion domain-specific pre-training is amplified when used on top of generic VLP. We see a similar trend for IRTF, where we see a gain of +2.5 on the Average metric when generic and domain-specific pre-training are coupled. FaD-VLP is competitive with state-of-the-art models on IRTF that use a comparable ResNet-50 visual encoder \cite{kim2021dual, zhao2022progressive}. \citet{zhao2022progressive} is a concurrent work on multistage vision-language pre-training; while it focuses only on the IRTF task, it shows gains of fashion-specific pre-training on top of a CLIP visual and textual encoder, and also sees benefits in scaling up to the larger ViT encoder. Additionally, we illustrate the effectiveness of FaD-VLP on another understanding task: multimodal categorization. Results on CR and SR indicate that the global fused representations are better for predicting categories and fine-grained subcategories after domain-specific pre-training.

\begin{figure*}
    \centering
    \includegraphics[width=\textwidth]{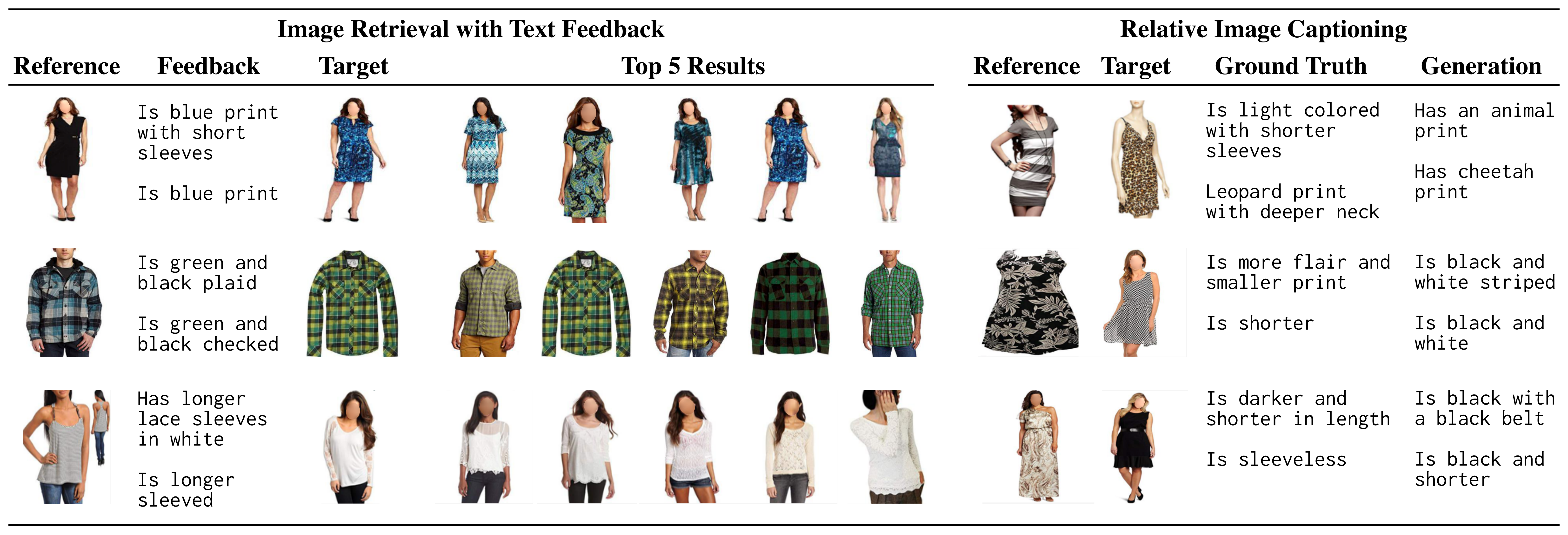}
    \caption{(Left) Examples of the top 5 results for three examples from FashionIQ on IRTF. For each reference image in FashionIQ, there are two pieces of text feedback. (Right) Examples of generations for three reference-target pairs in the RIC task.}
    \label{fig:qualitative-results}
\end{figure*}

Previous methods that evaluated IC were encoder-based: while these models, \textit{e.g.}, \cite{zhuge2021kaleido}, are evaluated on both ITR/TIR and IC, the IC performance was suboptimal because these methods used sequential Masked Language Modeling predictions for generating text at inference time. Our results highlight the benefit of our decoder-based architecture (even without CLIP initialization or domain-specific pre-training), which is effective at ITR/TIR \textit{and} IC. We see additional gains with generic VLP and domain-specific VLP. For RIC, we compare to a decoder baseline that concatenates the two input image representations and uses a single cross attention mechanism (with a CLIP initialization on the visual encoder). The RIC results further highlight the effectiveness of triplet pre-training on top of paired pre-training and CLIP initialization.

\subsection{Effect of Pre-training Tasks}
\label{sec:effect-of-pre-training-tasks}

We conduct a thorough ablation study on the pre-training tasks to analyze the impact of our various pre-training objectives (Table \ref{tab:ablation-pre-training}).

\paragraph{Effect of Paired Pre-training.} Our paired pre-training tasks (CMC and ICLM) provide a clear stepwise improvement on IRTF, ITR, TIR, and RIC (see lines (1), (2) and (6)-(8)), while the individual gains are less evident in the IC, CR, and SR tasks.
\paragraph{Effect of Triplet Pre-training.} Although the triplet pre-training tasks (HMC and RCLM) use no additional data, we find that they provide further improvement on top of the paired pre-training tasks (see lines (2)-(4) and (8)-(10)). This occurs in IRTF, where the triplet tasks lead to approximately 2 point gain, as well as ITR and TIR, where they lead to a 3 point gain (when used on top of the CLIP initialization). RIC also sees a gain with triplet pre-training of above 2.3 points. There appear to be smaller gains for IC, CR, and SR.
\paragraph{Effect of Bootstrapping.} We additionally experiment specifically with the effect of bootstrapping in the generation of relative captions (in which we feed samples from the multimodal decoder in the Relative Captioner mode to the HMC task in the Fuser mode). We find that bootstrapping yields a small benefit for IRTF, ITR, and TIR (see line (5) and (11)), indicating that the global representations may become more robust to noise when the Fuser is trained with more diverse relative captions.

\subsection{Qualitative Results}
We visualize results on IRTF and RIC in Figure~\ref{fig:qualitative-results}. These examples illustrate the trained model's ability to handle some compositional changes (\textit{e.g.}, adding a blue print as well as short sleeves). They also illustrate the model's ability to handle changes to certain mentioned attributes while not changing unmentioned attributes; for example, adding a blue print and short sleeves does not alter the length of the dress. Retrieving the target item in the top results is still challenging since the text may not sufficiently describe the target image (\textit{e.g.}, there are multiple white tops with lace sleeves) and because of the fine-grainedness of the changes described (\textit{e.g.}, only the sleeves should be lace). The RIC task is similarly challenging because there are multiple axes along which two images can differ, and often the ground truth may describe only one (\textit{e.g.}, ``is sleeveless''). Compared to the ground truth relative captions, the model tends to generate more concrete relative captions (\textit{e.g.}, involving color or relative length) rather than less concrete descriptors (\textit{e.g.}, ``more flair'').

\section{Conclusion}
\label{sec:conclusion}
Our work introduces a novel fashion-specific pre-training framework based on weakly-supervised triplets, constructed from paired fashion image-text data, and flexible decoder-based model architecture capable of both retrieval and captioning tasks in fashion. 
Our approach outperforms baselines on a diverse set of fashion tasks and highlights the value of fashion-specific pre-training, as well as the promise of triplet-based pre-training.

\section*{Limitations}
\label{sec:limitations}
Although fashion is a huge global industry, our experiments have been limited primarily to Western fashion styles with English language descriptions. It is likely that other clothing types contain properties that affect the performance of our method (e.g., different categories of attributes and different degrees of variation between products) and that other languages are less amenable to our token-wise method for constructing relative captions. Additionally, while this system sees gains from pre-training, corresponding limitations arise as well: we require several GPUs to pre-train FaD-VLP and to fine-tune it on our downstream tasks. Building smaller, more resource-efficient models with the same performance is an open question. We discuss broader limitations in the following section.

\section*{Broader Impact}
\label{sec:ethics-statement}
Our model is motivated by e-commerce applications, and demonstrates good performance on benchmarks, but it is not suitable for deployment or commercial use in its current form. It likely carries biases present in its training data: the texts in its pre-training dataset were sourced from human-provided descriptions, the human models are from shopping catalogues, and the distribution of fashion items in the datasets reflect societal stereotypes and expectations about characteristics such as gender. The technology presented in this paper could be used to support interactive shopping assistants, which have the potential to help users locate items that match their preferences. However, style is a personal, subjective form of self-expression, and so further research could focus on individualizing results to particular users.

\section*{Acknowledgements}

We thank the anonymous reviewers for their constructive feedback. We also thank Arka Sadhu, Brandon Han, Adina Williams, Alex Tamkin, Siddharth Karamcheti, and Benjamin Newman for helpful discussions. Fashion images in Figures \ref{fig:front-figure}, \ref{fig:pseudo-triplets}, and \ref{fig:qualitative-results} are from the data sources described in Section \ref{sec:datasets}.

\bibliography{references/anthology,references/custom}

\begin{thebibliography}{57}
\expandafter\ifx\csname natexlab\endcsname\relax\def\natexlab#1{#1}\fi

\bibitem[{Ak et~al.(2018)Ak, Lim, Tham, and Kassim}]{ak2018efficient}
Kenan~E. Ak, Joo{-}Hwee Lim, Jo~Yew Tham, and Ashraf~A. Kassim. 2018.
\newblock \href {https://doi.org/10.1109/WACV.2018.00186} {Efficient
  multi-attribute similarity learning towards attribute-based fashion search}.
\newblock In \emph{{IEEE} Winter Conference on Applications of Computer Vision
  (WACV)}.

\bibitem[{Banerjee and Lavie(2005)}]{banerjee2005meteor}
Satanjeev Banerjee and Alon Lavie. 2005.
\newblock \href {https://aclanthology.org/W05-0909/} {{METEOR:} {A}n automatic
  metric for {MT} evaluation with improved correlation with human judgments}.
\newblock In \emph{Proceedings of the Workshop on Intrinsic and Extrinsic
  Evaluation Measures for Machine Translation and/or Summarization@ACL}.

\bibitem[{Bird et~al.(2009)Bird, Klein, and Loper}]{bird2009natural}
Steven Bird, Ewan Klein, and Edward Loper. 2009.
\newblock \emph{Natural language processing with Python: analyzing text with
  the natural language toolkit}.
\newblock O'Reilly Media, Inc.

\bibitem[{Chen et~al.(2020{\natexlab{a}})Chen, Gong, and
  Bazzani}]{chen2020image}
Yanbei Chen, Shaogang Gong, and Loris Bazzani. 2020{\natexlab{a}}.
\newblock \href
  {https://openaccess.thecvf.com/content\_CVPR\_2020/html/Chen\_Image\_Search\_With\_Text\_Feedback\_by\_Visiolinguistic\_Attention\_Learning\_CVPR\_2020\_paper.html}
  {Image search with text feedback by visiolinguistic attention learning}.
\newblock In \emph{{IEEE/CVF} Conference on Computer Vision and Pattern
  Recognition (CVPR)}.

\bibitem[{Chen et~al.(2020{\natexlab{b}})Chen, Li, Yu, Kholy, Ahmed, Gan,
  Cheng, and Liu}]{chen2020uniter}
Yen{-}Chun Chen, Linjie Li, Licheng Yu, Ahmed~El Kholy, Faisal Ahmed, Zhe Gan,
  Yu~Cheng, and Jingjing Liu. 2020{\natexlab{b}}.
\newblock \href {https://doi.org/10.1007/978-3-030-58577-8\_7} {{UNITER:}
  {U}niversal image-text representation learning}.
\newblock In \emph{European Conference on Computer Vision ({ECCV})}.

\bibitem[{Deldjoo et~al.(2022)Deldjoo, Nazary, Ramisa, McAuley, Pellegrini,
  Bellog{\'{\i}}n, and Noia}]{deldjoo2022review}
Yashar Deldjoo, Fatemeh Nazary, Arnau Ramisa, Julian~J. McAuley, Giovanni
  Pellegrini, Alejandro Bellog{\'{\i}}n, and Tommaso~Di Noia. 2022.
\newblock \href {http://arxiv.org/abs/2202.02757} {A review of modern fashion
  recommender systems}.
\newblock \emph{arXiv}, abs/2202.02757.

\bibitem[{Devlin et~al.(2019)Devlin, Chang, Lee, and
  Toutanova}]{devlin2019bert}
Jacob Devlin, Ming{-}Wei Chang, Kenton Lee, and Kristina Toutanova. 2019.
\newblock \href {https://doi.org/10.18653/v1/n19-1423} {{BERT:} pre-training of
  deep bidirectional transformers for language understanding}.
\newblock In \emph{Conference of the North American Chapter of the Association
  for Computational Linguistics: Human Language Technologies, {(NAACL-HLT)}}.

\bibitem[{Dong et~al.(2021)Dong, Zhan, Wu, Wei, Wei, Lu, and
  Liang}]{dong2021m5product}
Xiao Dong, Xunlin Zhan, Yangxin Wu, Yunchao Wei, Xiaoyong Wei, Minlong Lu, and
  Xiaodan Liang. 2021.
\newblock \href {http://arxiv.org/abs/2109.04275} {{M5Product}: {A} multi-modal
  pretraining benchmark for e-commercial product downstream tasks}.
\newblock \emph{arXiv}, abs/2109.04275.

\bibitem[{Faghri et~al.(2018)Faghri, Fleet, Kiros, and
  Fidler}]{faghri2018vsepp}
Fartash Faghri, David~J. Fleet, Jamie~Ryan Kiros, and Sanja Fidler. 2018.
\newblock \href {http://bmvc2018.org/contents/papers/0344.pdf} {{VSE++:}
  {I}mproving visual-semantic embeddings with hard negatives}.
\newblock In \emph{British Machine Vision Conference 2018 ({BMVC})}.

\bibitem[{Gao et~al.(2020)Gao, Jin, Chen, Qiu, Li, Wei, Hu, and
  Wang}]{gao2020fashionbert}
Dehong Gao, Linbo Jin, Ben Chen, Minghui Qiu, Peng Li, Yi~Wei, Yi~Hu, and Hao
  Wang. 2020.
\newblock \href {https://doi.org/10.1145/3397271.3401430} {{FashionBERT}:
  {T}ext and image matching with adaptive loss for cross-modal retrieval}.
\newblock In \emph{{ACM} {SIGIR} Conference on Research and Development in
  Information Retrieval}.

\bibitem[{Goenka et~al.(2022)Goenka, Zheng, Jaiswal, Chada, Wu, Hedau, and
  Natarajan}]{goenka2022fashionvlp}
Sonam Goenka, Zhaoheng Zheng, Ayush Jaiswal, Rakesh Chada, Yue Wu, Varsha
  Hedau, and Pradeep Natarajan. 2022.
\newblock \href
  {https://openaccess.thecvf.com/content/CVPR2022/papers/Goenka_FashionVLP_Vision_Language_Transformer_for_Fashion_Retrieval_With_Feedback_CVPR_2022_paper.pdf}
  {{FashionVLP}: {V}ision language transformer for fashion retrieval with
  feedback}.
\newblock In \emph{IEEE/CVF Conference on Computer Vision and Pattern
  Recognition ({CVPR})}.

\bibitem[{Han et~al.(2022)Han, He, Zhang, Song, and Xiang}]{han2022uigr}
Xiao Han, Sen He, Li~Zhang, Yi{-}Zhe Song, and Tao Xiang. 2022.
\newblock \href {https://doi.org/10.48550/arXiv.2204.03111} {{UIGR:} {U}nified
  interactive garment retrieval}.
\newblock \emph{arXiv}, abs/2204.03111.

\bibitem[{Han et~al.(2017)Han, Wu, Huang, Zhang, Zhu, Li, Zhao, and
  Davis}]{han2017automatic}
Xintong Han, Zuxuan Wu, Phoenix~X. Huang, Xiao Zhang, Menglong Zhu, Yuan Li,
  Yang Zhao, and Larry~S. Davis. 2017.
\newblock \href {https://doi.org/10.1109/ICCV.2017.163} {Automatic
  spatially-aware fashion concept discovery}.
\newblock In \emph{International Conference on Computer Vision ({ICCV})}.

\bibitem[{Han et~al.(2018)Han, Wu, Wu, Yu, and Davis}]{han2018viton}
Xintong Han, Zuxuan Wu, Zhe Wu, Ruichi Yu, and Larry~S. Davis. 2018.
\newblock \href
  {http://openaccess.thecvf.com/content\_cvpr\_2018/html/Han\_VITON\_An\_Image-Based\_CVPR\_2018\_paper.html}
  {{VITON:} {A}n image-based virtual try-on network}.
\newblock In \emph{{IEEE} Conference on Computer Vision and Pattern Recognition
  ({CVPR})}.

\bibitem[{He et~al.(2016)He, Zhang, Ren, and Sun}]{he2016deep}
Kaiming He, Xiangyu Zhang, Shaoqing Ren, and Jian Sun. 2016.
\newblock \href {https://doi.org/10.1109/CVPR.2016.90} {Deep residual learning
  for image recognition}.
\newblock In \emph{2016 {IEEE} Conference on Computer Vision and Pattern
  Recognition, {CVPR} 2016, Las Vegas, NV, USA, June 27-30, 2016}. {IEEE}
  Computer Society.

\bibitem[{Jia et~al.(2021)Jia, Yang, Xia, Chen, Parekh, Pham, Le, Sung, Li, and
  Duerig}]{jia2021align}
Chao Jia, Yinfei Yang, Ye~Xia, Yi{-}Ting Chen, Zarana Parekh, Hieu Pham,
  Quoc~V. Le, Yun{-}Hsuan Sung, Zhen Li, and Tom Duerig. 2021.
\newblock \href {http://proceedings.mlr.press/v139/jia21b.html} {Scaling up
  visual and vision-language representation learning with noisy text
  supervision}.
\newblock In \emph{International Conference on Machine Learning ({ICML})}.

\bibitem[{Kim et~al.(2021)Kim, Yu, Kim, and Kim}]{kim2021dual}
Jongseok Kim, Youngjae Yu, Hoeseong Kim, and Gunhee Kim. 2021.
\newblock \href {https://ojs.aaai.org/index.php/AAAI/article/view/16271} {Dual
  compositional learning in interactive image retrieval}.
\newblock In \emph{{AAAI} Conference on Artificial Intelligence}.

\bibitem[{Kiros et~al.(2014)Kiros, Salakhutdinov, and
  Zemel}]{kiros2014unifying}
Ryan Kiros, Ruslan Salakhutdinov, and Richard~S. Zemel. 2014.
\newblock \href {http://arxiv.org/abs/1411.2539} {Unifying visual-semantic
  embeddings with multimodal neural language models}.
\newblock \emph{arXiv}, abs/1411.2539.

\bibitem[{Lee et~al.(2018)Lee, Chen, Hua, Hu, and He}]{lee2018stacked}
Kuang{-}Huei Lee, Xi~Chen, Gang Hua, Houdong Hu, and Xiaodong He. 2018.
\newblock \href {https://doi.org/10.1007/978-3-030-01225-0\_13} {Stacked cross
  attention for image-text matching}.
\newblock In \emph{European Conference on Computer Vision ({ECCV})}, volume
  11208.

\bibitem[{Lee et~al.(2021)Lee, Kim, and Han}]{lee2021cosmo}
Seungmin Lee, Dongwan Kim, and Bohyung Han. 2021.
\newblock \href
  {https://openaccess.thecvf.com/content/CVPR2021/html/Lee\_CoSMo\_Content-Style\_Modulation\_for\_Image\_Retrieval\_With\_Text\_Feedback\_CVPR\_2021\_paper.html}
  {{CoSMo}: {C}ontent-style modulation for image retrieval with text feedback}.
\newblock In \emph{{IEEE} Conference on Computer Vision and Pattern Recognition
  ({CVPR})}.

\bibitem[{Li et~al.(2022)Li, Li, Xiong, and Hoi}]{li2022blip}
Junnan Li, Dongxu Li, Caiming Xiong, and Steven C.~H. Hoi. 2022.
\newblock \href {https://arxiv.org/abs/2201.12086} {{BLIP:} {B}ootstrapping
  language-image pre-training for unified vision-language understanding and
  generation}.
\newblock In \emph{International Conference on Machine Learning (ICML)}.

\bibitem[{Li et~al.(2021{\natexlab{a}})Li, Selvaraju, Gotmare, Joty, Xiong, and
  Hoi}]{li2021align}
Junnan Li, Ramprasaath~R. Selvaraju, Akhilesh Gotmare, Shafiq~R. Joty, Caiming
  Xiong, and Steven~Chu{-}Hong Hoi. 2021{\natexlab{a}}.
\newblock \href
  {https://proceedings.neurips.cc/paper/2021/hash/505259756244493872b7709a8a01b536-Abstract.html}
  {Align before fuse: Vision and language representation learning with momentum
  distillation}.
\newblock In \emph{Advances in Neural Information Processing Systems
  (NeurIPS)}.

\bibitem[{Li et~al.(2019)Li, Yatskar, Yin, Hsieh, and Chang}]{li2019visualbert}
Liunian~Harold Li, Mark Yatskar, Da~Yin, Cho{-}Jui Hsieh, and Kai{-}Wei Chang.
  2019.
\newblock \href {http://arxiv.org/abs/1908.03557} {{VisualBERT}: {A} simple and
  performant baseline for vision and language}.
\newblock \emph{arXiv}, abs/1908.03557.

\bibitem[{Li et~al.(2021{\natexlab{b}})Li, Gao, Niu, Xiao, Liu, Liu, Wu, and
  Wang}]{li2021unimo}
Wei Li, Can Gao, Guocheng Niu, Xinyan Xiao, Hao Liu, Jiachen Liu, Hua Wu, and
  Haifeng Wang. 2021{\natexlab{b}}.
\newblock \href {https://doi.org/10.18653/v1/2021.acl-long.202} {{UNIMO:}
  towards unified-modal understanding and generation via cross-modal
  contrastive learning}.
\newblock In \emph{Annual Meeting of the Association for Computational
  Linguistics and the 11th International Joint Conference on Natural Language
  Processing ({ACL/IJCNLP})}.

\bibitem[{Li et~al.(2020)Li, Yin, Li, Zhang, Hu, Zhang, Wang, Hu, Dong, Wei,
  Choi, and Gao}]{li2020oscar}
Xiujun Li, Xi~Yin, Chunyuan Li, Pengchuan Zhang, Xiaowei Hu, Lei Zhang, Lijuan
  Wang, Houdong Hu, Li~Dong, Furu Wei, Yejin Choi, and Jianfeng Gao. 2020.
\newblock \href {https://doi.org/10.1007/978-3-030-58577-8\_8} {Oscar:
  Object-semantics aligned pre-training for vision-language tasks}.
\newblock In \emph{European Conference on Computer Vision ({ECCV})}, volume
  12375.

\bibitem[{Lin(2004)}]{lin2004rouge}
Chin-Yew Lin. 2004.
\newblock {ROUGE}: {A} package for automatic evaluation of summaries.
\newblock In \emph{Text Summarization Branches Out: Proceedings of the ACL-04
  Workshop}.

\bibitem[{Liu et~al.(2021)Liu, Opazo, Teney, and Gould}]{liu2021image}
Zheyuan Liu, Cristian~Rodriguez Opazo, Damien Teney, and Stephen Gould. 2021.
\newblock \href {https://doi.org/10.1109/ICCV48922.2021.00213} {Image retrieval
  on real-life images with pre-trained vision-and-language models}.
\newblock In \emph{2021 {IEEE/CVF} International Conference on Computer Vision
  ({ICCV})}. {IEEE}.

\bibitem[{Liu et~al.(2016)Liu, Luo, Qiu, Wang, and Tang}]{liu2016deep}
Ziwei Liu, Ping Luo, Shi Qiu, Xiaogang Wang, and Xiaoou Tang. 2016.
\newblock \href {https://doi.org/10.1109/CVPR.2016.124} {{DeepFashion}:
  {P}owering robust clothes recognition and retrieval with rich annotations}.
\newblock In \emph{{IEEE} Conference on Computer Vision and Pattern Recognition
  (CVPR)}.

\bibitem[{Lu et~al.(2019)Lu, Batra, Parikh, and Lee}]{lu2019vilbert}
Jiasen Lu, Dhruv Batra, Devi Parikh, and Stefan Lee. 2019.
\newblock \href
  {https://proceedings.neurips.cc/paper/2019/hash/c74d97b01eae257e44aa9d5bade97baf-Abstract.html}
  {{ViLBERT}: Pretraining task-agnostic visiolinguistic representations for
  vision-and-language tasks}.
\newblock In \emph{Advances in Neural Information Processing Systems
  ({NeurIPS})}.

\bibitem[{McAuley et~al.(2015)McAuley, Targett, Shi, and van~den
  Hengel}]{mcauley2015image}
Julian~J. McAuley, Christopher Targett, Qinfeng Shi, and Anton van~den Hengel.
  2015.
\newblock \href {https://doi.org/10.1145/2766462.2767755} {Image-based
  recommendations on styles and substitutes}.
\newblock In \emph{International {ACM} Conference on Research and Development
  in Information Retrieval ({SIGIR})}.

\bibitem[{Papineni et~al.(2002)Papineni, Roukos, Ward, and
  Zhu}]{papineni2002bleu}
Kishore Papineni, Salim Roukos, Todd Ward, and Wei{-}Jing Zhu. 2002.
\newblock \href {https://aclanthology.org/P02-1040/} {Bleu: a method for
  automatic evaluation of machine translation}.
\newblock In \emph{Association for Computational Linguistics ({ACL})}. {ACL}.

\bibitem[{Paszke et~al.(2019)Paszke, Gross, Massa, Lerer, Bradbury, Chanan,
  Killeen, Lin, Gimelshein, Antiga, Desmaison, K{\"{o}}pf, Yang, DeVito,
  Raison, Tejani, Chilamkurthy, Steiner, Fang, Bai, and
  Chintala}]{paszke2019pytorch}
Adam Paszke, Sam Gross, Francisco Massa, Adam Lerer, James Bradbury, Gregory
  Chanan, Trevor Killeen, Zeming Lin, Natalia Gimelshein, Luca Antiga, Alban
  Desmaison, Andreas K{\"{o}}pf, Edward~Z. Yang, Zachary DeVito, Martin Raison,
  Alykhan Tejani, Sasank Chilamkurthy, Benoit Steiner, Lu~Fang, Junjie Bai, and
  Soumith Chintala. 2019.
\newblock \href
  {https://proceedings.neurips.cc/paper/2019/hash/bdbca288fee7f92f2bfa9f7012727740-Abstract.html}
  {{PyTorch}: {A}n imperative style, high-performance deep learning library}.
\newblock In \emph{Advances in Neural Information Processing Systems
  ({NeurIPS})}.

\bibitem[{Qi et~al.(2020)Qi, Su, Song, Cui, Bharti, and
  Sacheti}]{qi2020imagebert}
Di~Qi, Lin Su, Jia Song, Edward Cui, Taroon Bharti, and Arun Sacheti. 2020.
\newblock \href {http://arxiv.org/abs/2001.07966} {{ImageBERT}: Cross-modal
  pre-training with large-scale weak-supervised image-text data}.
\newblock \emph{arXiv}, abs/2001.07966.

\bibitem[{Radford et~al.(2021{\natexlab{a}})Radford, Kim, Hallacy, Ramesh, Goh,
  Agarwal, Sastry, Askell, Mishkin, Clark, Krueger, and
  Sutskever}]{radford2021clip}
Alec Radford, Jong~Wook Kim, Chris Hallacy, Aditya Ramesh, Gabriel Goh,
  Sandhini Agarwal, Girish Sastry, Amanda Askell, Pamela Mishkin, Jack Clark,
  Gretchen Krueger, and Ilya Sutskever. 2021{\natexlab{a}}.
\newblock \href {http://proceedings.mlr.press/v139/radford21a.html} {Learning
  transferable visual models from natural language supervision}.
\newblock In \emph{International Conference on Machine Learning ({ICML})}.

\bibitem[{Radford et~al.(2021{\natexlab{b}})Radford, Kim, Hallacy, Ramesh, Goh,
  Agarwal, Sastry, Askell, Mishkin, Clark, Krueger, and
  Sutskever}]{radford2021learning}
Alec Radford, Jong~Wook Kim, Chris Hallacy, Aditya Ramesh, Gabriel Goh,
  Sandhini Agarwal, Girish Sastry, Amanda Askell, Pamela Mishkin, Jack Clark,
  Gretchen Krueger, and Ilya Sutskever. 2021{\natexlab{b}}.
\newblock \href {http://proceedings.mlr.press/v139/radford21a.html} {Learning
  transferable visual models from natural language supervision}.
\newblock In \emph{International Conference on Machine Learning (ICML)}.

\bibitem[{Reimers and Gurevych(2019)}]{reimers2019sentence}
Nils Reimers and Iryna Gurevych. 2019.
\newblock \href {https://arxiv.org/abs/1908.10084} {{Sentence-BERT}: Sentence
  embeddings using siamese bert-networks}.
\newblock In \emph{Conference on Empirical Methods in Natural Language
  Processing ({EMNLP})}.

\bibitem[{Rostamzadeh et~al.(2018)Rostamzadeh, Hosseini, Boquet, Stokowiec,
  Zhang, Jauvin, and Pal}]{rostamzadeh2018fashion}
Negar Rostamzadeh, Seyedarian Hosseini, Thomas Boquet, Wojciech Stokowiec, Ying
  Zhang, Christian Jauvin, and Chris Pal. 2018.
\newblock \href {http://arxiv.org/abs/1806.08317} {{Fashion-Gen}: {T}he
  generative fashion dataset and challenge}.
\newblock \emph{ICML Workshop on Theoretical Foundations and Applications of
  Deep Generative Models}.

\bibitem[{Su et~al.(2020)Su, Zhu, Cao, Li, Lu, Wei, and Dai}]{su2020vlbert}
Weijie Su, Xizhou Zhu, Yue Cao, Bin Li, Lewei Lu, Furu Wei, and Jifeng Dai.
  2020.
\newblock \href {https://arxiv.org/abs/1908.08530} {{VL-BERT:} {P}re-training
  of generic visual-linguistic representations}.
\newblock In \emph{International Conference on Learning Representations
  ({ICLR})}.

\bibitem[{Tan and Bansal(2019)}]{tan2019lxmert}
Hao Tan and Mohit Bansal. 2019.
\newblock \href {https://doi.org/10.18653/v1/D19-1514} {{LXMERT:} {L}earning
  cross-modality encoder representations from transformers}.
\newblock In \emph{Conference on Empirical Methods in Natural Language
  Processing and the 9th International Joint Conference on Natural Language
  Processing ({EMNLP-IJCNLP})}.

\bibitem[{van~den Oord et~al.(2018)van~den Oord, Li, and
  Vinyals}]{oord2018representation}
A{\"{a}}ron van~den Oord, Yazhe Li, and Oriol Vinyals. 2018.
\newblock \href {http://arxiv.org/abs/1807.03748} {Representation learning with
  contrastive predictive coding}.
\newblock In \emph{Advances in Neural Information Processing Systems ({NIPS})}.

\bibitem[{Vaswani et~al.(2017)Vaswani, Shazeer, Parmar, Uszkoreit, Jones,
  Gomez, Kaiser, and Polosukhin}]{vaswani2017transformer}
Ashish Vaswani, Noam Shazeer, Niki Parmar, Jakob Uszkoreit, Llion Jones,
  Aidan~N. Gomez, Lukasz Kaiser, and Illia Polosukhin. 2017.
\newblock \href
  {https://proceedings.neurips.cc/paper/2017/hash/3f5ee243547dee91fbd053c1c4a845aa-Abstract.html}
  {Attention is all you need}.
\newblock In \emph{Advances in Neural Information Processing ({NeurIPS})}.

\bibitem[{Vedantam et~al.(2015)Vedantam, Zitnick, and
  Parikh}]{vedantam205cider}
Ramakrishna Vedantam, C.~Lawrence Zitnick, and Devi Parikh. 2015.
\newblock \href {https://doi.org/10.1109/CVPR.2015.7299087} {{CIDEr}:
  {C}onsensus-based image description evaluation}.
\newblock In \emph{{IEEE} Conference on Computer Vision and Pattern Recognition
  (CVPR)}.

\bibitem[{Vo et~al.(2019)Vo, Jiang, Sun, Murphy, Li, Fei{-}Fei, and
  Hays}]{vo2019composing}
Nam Vo, Lu~Jiang, Chen Sun, Kevin Murphy, Li{-}Jia Li, Li~Fei{-}Fei, and James
  Hays. 2019.
\newblock \href
  {http://openaccess.thecvf.com/content\_CVPR\_2019/html/Vo\_Composing\_Text\_and\_Image\_for\_Image\_Retrieval\_-\_an\_Empirical\_CVPR\_2019\_paper.html}
  {Composing text and image for image retrieval - an empirical odyssey}.
\newblock In \emph{{IEEE} Conference on Computer Vision and Pattern Recognition
  ({CVPR})}.

\bibitem[{Wang et~al.(2019)Wang, Yang, Qian, Ma, Lu, Li, and
  Fan}]{wang2019position}
Yaxiong Wang, Hao Yang, Xueming Qian, Lin Ma, Jing Lu, Biao Li, and Xin Fan.
  2019.
\newblock \href {https://doi.org/10.24963/ijcai.2019/526} {Position focused
  attention network for image-text matching}.
\newblock In \emph{International Joint Conference on Intelligence ({IJCAI})}.

\bibitem[{Wang et~al.(2022)Wang, Yu, Yu, Dai, Tsvetkov, and
  Cao}]{wang2021simvlm}
Zirui Wang, Jiahui Yu, Adams~Wei Yu, Zihang Dai, Yulia Tsvetkov, and Yuan Cao.
  2022.
\newblock \href {https://arxiv.org/abs/2108.10904} {{SimVLM}: {S}imple visual
  language model pretraining with weak supervision}.
\newblock In \emph{International Conference on Learning Representations
  ({ICLR})}.

\bibitem[{Wu et~al.(2021)Wu, Gao, Guo, Al{-}Halah, Rennie, Grauman, and
  Feris}]{wu2021fashion}
Hui Wu, Yupeng Gao, Xiaoxiao Guo, Ziad Al{-}Halah, Steven Rennie, Kristen
  Grauman, and Rog{\'{e}}rio Feris. 2021.
\newblock \href
  {https://openaccess.thecvf.com/content/CVPR2021/html/Wu\_Fashion\_IQ\_A\_New\_Dataset\_Towards\_Retrieving\_Images\_by\_Natural\_CVPR\_2021\_paper.html}
  {Fashion {IQ:} {A} new dataset towards retrieving images by natural language
  feedback}.
\newblock In \emph{{IEEE} Conference on Computer Vision and Pattern Recognition
  ({CVPR})}.

\bibitem[{Yang et~al.(2022)Yang, Yu, and Liu}]{yang2022full}
Han Yang, Xinrui Yu, and Ziwei Liu. 2022.
\newblock \href
  {https://openaccess.thecvf.com/content/CVPR2022/papers/Yang_Full-Range_Virtual_Try-On_With_Recurrent_Tri-Level_Transform_CVPR_2022_paper.pdf}
  {Full-range virtual try-on with recurrent tri-level transform}.
\newblock In \emph{IEEE/CVF Conference on Computer Vision and Pattern
  Recognition ({CVPR})}.

\bibitem[{Yang et~al.(2020)Yang, Zhang, Jin, Liu, Wu, Tan, Xie, Wang, and
  Wang}]{yang2020fashion}
Xuewen Yang, Heming Zhang, Di~Jin, Yingru Liu, Chi{-}Hao Wu, Jianchao Tan,
  Dongliang Xie, Jue Wang, and Xin Wang. 2020.
\newblock \href {https://doi.org/10.1007/978-3-030-58601-0\_1} {Fashion
  captioning: {T}owards generating accurate descriptions with semantic
  rewards}.
\newblock In \emph{European Conference on Computer Vision ({ECCV})}.

\bibitem[{Yu et~al.(2022{\natexlab{a}})Yu, Wang, Vasudevan, Yeung,
  Seyedhosseini, and Wu}]{yu2022coca}
Jiahui Yu, Zirui Wang, Vijay Vasudevan, Legg Yeung, Mojtaba Seyedhosseini, and
  Yonghui Wu. 2022{\natexlab{a}}.
\newblock \href {https://doi.org/10.48550/arXiv.2205.01917} {{CoCa}:
  {C}ontrastive captioners are image-text foundation models}.
\newblock \emph{arXiv}, abs/2205.01917.

\bibitem[{Yu et~al.(2022{\natexlab{b}})Yu, Chen, Sinha, Wang, Chen, Berg, and
  Zhang}]{yu2022commercemm}
Licheng Yu, Jun Chen, Animesh Sinha, Mengjiao M.~J. Wang, Hugo Chen, Tamara~L.
  Berg, and Ning Zhang. 2022{\natexlab{b}}.
\newblock \href {https://arxiv.org/abs/2202.07247} {{CommerceMM}: {L}arge-scale
  commerce multimodal representation learning with omni retrieval}.
\newblock \emph{arXiv}, abs/2202.07247.

\bibitem[{Yuan and Lam(2021)}]{yuan2021conversational}
Yifei Yuan and Wai Lam. 2021.
\newblock \href {https://doi.org/10.1145/3404835.3462881} {Conversational
  fashion image retrieval via multiturn natural language feedback}.
\newblock In \emph{International {ACM} Conference on Research and Development
  in Information Retrieval ({SIGIR})}.

\bibitem[{Zhang et~al.(2020)Zhang, Jiang, Miura, Manning, and
  Langlotz}]{zhang2020contrastive}
Yuhao Zhang, Hang Jiang, Yasuhide Miura, Christopher~D. Manning, and Curtis~P.
  Langlotz. 2020.
\newblock \href {http://arxiv.org/abs/2010.00747} {Contrastive learning of
  medical visual representations from paired images and text}.
\newblock \emph{arXiv}, abs/2010.00747.

\bibitem[{Zhang et~al.(2021)Zhang, Ma, Zhou, Men, Li, Ding, Tang, Zhou, and
  Yang}]{zhang2021ufcbert}
Zhu Zhang, Jianxin Ma, Chang Zhou, Rui Men, Zhikang Li, Ming Ding, Jie Tang,
  Jingren Zhou, and Hongxia Yang. 2021.
\newblock \href
  {https://proceedings.neurips.cc/paper/2021/hash/e46bc064f8e92ac2c404b9871b2a4ef2-Abstract.html}
  {{UFC-BERT:} unifying multi-modal controls for conditional image synthesis}.
\newblock In \emph{Advances in Neural Information Processing Systems
  ({NeurIPS})}.

\bibitem[{Zhao et~al.(2022)Zhao, Song, and Jin}]{zhao2022progressive}
Yida Zhao, Yuqing Song, and Qin Jin. 2022.
\newblock \href {https://doi.org/10.1145/3477495.3532047} {Progressive learning
  for image retrieval with hybrid-modality queries}.
\newblock In \emph{International {ACM} Conference on Research and Development
  in Information Retrieval ({SIGIR})}.

\bibitem[{Zhou et~al.(2020)Zhou, Palangi, Zhang, Hu, Corso, and
  Gao}]{zhou2020unifiedvlp}
Luowei Zhou, Hamid Palangi, Lei Zhang, Houdong Hu, Jason~J. Corso, and Jianfeng
  Gao. 2020.
\newblock \href {https://ojs.aaai.org/index.php/AAAI/article/view/7005}
  {Unified vision-language pre-training for image captioning and {VQA}}.
\newblock In \emph{{AAAI} Conference on Artificial Intelligence}.

\bibitem[{Zhu et~al.(2021)Zhu, Zhao, Zhang, Ye, Chen, Zhang, and
  Chen}]{zhu2021k3m}
Yushan Zhu, Huaixiao Zhao, Wen Zhang, Ganqiang Ye, Hui Chen, Ningyu Zhang, and
  Huajun Chen. 2021.
\newblock \href {https://doi.org/10.1145/3474085.3475648} {Knowledge perceived
  multi-modal pretraining in e-commerce}.
\newblock In \emph{{MM} '21: {ACM} Multimedia Conference}.

\bibitem[{Zhuge et~al.(2021)Zhuge, Gao, Fan, Jin, Chen, Zhou, Qiu, and
  Shao}]{zhuge2021kaleido}
Mingchen Zhuge, Dehong Gao, Deng{-}Ping Fan, Linbo Jin, Ben Chen, Haoming Zhou,
  Minghui Qiu, and Ling Shao. 2021.
\newblock \href
  {https://openaccess.thecvf.com/content/CVPR2021/html/Zhuge\_Kaleido-BERT\_Vision-Language\_Pre-Training\_on\_Fashion\_Domain\_CVPR\_2021\_paper.html}
  {{Kaleido-BERT}: {V}ision-language pre-training on fashion domain}.
\newblock In \emph{{IEEE} Conference on Computer Vision and Pattern Recognition
  ({CVPR})}.

\end{thebibliography}
\bibliographystyle{references/acl_natbib}

\appendix

\newpage
\section{Training Details}
\label{app:training-details}
This section describes our implementation and training details for pre-training and fine-tuning on each downstream task.

\paragraph{Implementation.} In our model, we use a ResNet-50 image encoder \cite{he2016deep} initialized from CLIP \cite{radford2021learning}. We initialize the text and multimodal decoder from BERT-base \cite{devlin2019bert}, using the first 6 layers for the text decoder and the second 6 layers for the multimodal decoder. The dimensionality of our joint embedding space is 2048.
We implement our models in PyTorch \cite{paszke2019pytorch} and pre-train on two 8 GPU NVIDIA A100 nodes.

\paragraph{Pre-training Datasets.} In the table below, we show statistics for the 5 datasets we use in our pre-training dataset. For datasets that are split into train and validation, we only use the training set.

\begin{table}[H]
\begin{small}
    \centering
    \begin{tabular}{lll}
\toprule
\textbf{Name} & \textbf{\# Pairs} & \textbf{\# Tokens} \\
\midrule
        FACAD &              888K &   21.15 $\pm$ 4.49 \\
  Fashion-Gen &              260K &  37.35 $\pm$ 14.70 \\
  Fashion200K &              172K &    4.84 $\pm$ 1.32 \\
 Shopping100K &              100K &   20.00 $\pm$ 3.99 \\
  DeepFashion &               26K &  53.03 $\pm$ 19.02 \\
\bottomrule
\end{tabular}
    \caption{Statistics for datasets in our pre-training dataset: the number of image-caption pairs and the mean $\pm$ std. number of tokens per caption. Token statistics are calculated on a sample of 10K entries.}
    \label{tab:appendix-dataset_statistics}
\end{small}
\end{table}

\noindent 
Below, we describe each dataset briefly:

\noindent \textbf{FACAD} \cite{yang2020fashion} is a dataset of fashion images and descriptions crawled from the web, with 888K pairs in its training set.

\noindent \textbf{DeepFashion} \cite{liu2016deep} consists of diverse clothing images with different annotation types. We use the 26K images and descriptions in the In-shop Clothes Retrieval benchmark, which is sourced from an online catalogue. We use a concatenation of the color annotation and in-shop description as the caption. 

\noindent \textbf{Shopping100k} \cite{ak2018efficient} consists of 100K pairs of fashion images and attributes that are collected from meta-data on shopping websites. Each image has at least 5 structured attributes, which we concatenate to form a caption (\textit{e.g.}, \textit{t-shirt with white color and jersey fabric and regular fit}).

\noindent \textbf{Fashion-Gen} \cite{rostamzadeh2018fashion} consists of 260K fashion images and detailed descriptions, annotated by professional stylists, in its training set. The image span 48 categories and 121 fine-grained subcategories.

\noindent \textbf{Fashion200K} \cite{han2017automatic} consists of 172K clothing images in its training set along with descriptions from shopping websites. We exclude 15 pairs (0.0087\%) that have broken image links. We use a concatenation of the attribute annotations as the caption.

After assembling pre-training dataset, we hold out a random split of $3\%$ of the data (for any validation needed on pre-training data). Our final pre-training dataset has 1.4 million image-text pairs.

\paragraph{Pre-training Details.} We have two pre-training stages, one for paired pre-training and the second for paired pre-training plus triplet pre-training; we found the two-stage setup to be more effective than a single stage. For both stages, we sum the appropriate losses. We can calculate CMC/ICLM from a single image-text pair and HMC/RCLM from a single triplet (with the same reference image as the pair). For pre-training, we use the Adam optimizer with a learning rate of $10^{-5}$. We use a batch size of 512 split over 2 nodes with a total of 16 GPUs. In the first stage of pre-training (paired pre-training tasks), we train for 82K steps, taking approximately 1.5 days. In the optional second stage (paired pre-training plus triplet pre-training tasks), we train for 55K steps, which takes another approximately 1.5 days. Our model has $1.9 \times 10^8$ parameters. We use the following preprocessing/data augmentations on the images: resize to $224\times224$ pixels with a random crop, that has random scaling between 0.8 and 1.0 and aspect ratio between 0.75 and 1.3, along with a random horizontal flip. We use the same preprocessing for downstream fine-tuning.

For constructed triplets, we use $|S|=1000$, and we weight the three metrics in $\Delta$ to have approximately equal relative weight by setting set $\lambda_1=1$, $\lambda_2=1$, and $\lambda_3$ = $\frac{1}{16}$. We use $\texttt{nltk}$ \cite{bird2009natural} for word and sentence tokenization. We sample from the following templates when constructing relative captions: \{\texttt{modify <r> to be <t>}, \texttt{<t> instead of <r>}, \texttt{change <r> to <t>}, and \texttt{replace <r> with <t>}\} where \texttt{<r>} and \texttt{<t>} represent the reference and target tokens respectively from the procedure from Section \ref{sec:constructing-pseudo-triplets}. If either \texttt{<r>} or \texttt{<t>} is empty, we run the procedure again on the next sentence in the reference and target captions.

\section{Fine-tuning Details}

We provide details on our fine-tuning stage for each downstream task below. For all tasks, we use a batch size of 128 distributed over 4 GPUs.

\paragraph{ITR/TIR.} We use the Adam optimizer with a learning rate of $10^{-5}$ and fine-tune for up to $120$K steps. Fine-tuning takes approximately 20 hours.

\paragraph{IRTF.} We use the Adam optimizer with a warmup learning rate of $10^{-6}$ for 140 steps, followed cosine decay from $3\times10^{-5}$ over 4,100 steps.  Fine-tuning takes about 1.5 hours. In our ablation, we found that models without pre-training take longer to converge for this task; to strengthen our baseline comparison, we train these for 25K steps, which takes approximately 5 hours.

\paragraph{CR/SR.} We use the Adam optimizer with a learning rate of $10^{-5}$ and fine-tune for up to 81K steps. Fine-tuning takes approximately 14 hours.

\paragraph{IC.} We use the Adam optimizer with a learning rate of $10^{-5}$ and fine-tune for up to 40K steps. Fine-tuning takes approximately 7 hours.

\paragraph{RIC.} We use the Adam optimizer with a learning rate of $10^{-5}$ and fine-tune for up to 3K steps. Fine-tuning takes approximately 1 hour.
\section{Evaluation Details}

This section describes our evaluation pipeline for the 7 downstream tasks.

\paragraph{ITR/TIR.} We evaluate cross-modal retrieval on Fashion-Gen \cite{rostamzadeh2018fashion}. We compare to the results reported by \citet{zhuge2021kaleido} and for fair comparison, follow their evaluation protocol, which works as follows. For each image (or text) query in the evaluation set of 32,528 queries, the model is to pick the matched text (or image) out of a sample of 101 candidates. The 101 candidates contain 1 correct match as well as 100 sampled texts (or images) from other products in the same subcategory. If a subcategory has less than 100 products, we sample the remaining negative examples from the same category. Since the random sample of candidates for each query is not released by the authors, we generate 5 random samples of candidates for our experiments and report the average. Like \citet{zhuge2021kaleido}, we report Recall@1, Recall@5, and Recall@10.

\paragraph{IRTF.} We evaluate this task on Fashion IQ \cite{wu2021fashion} using the protocol laid out by the authors and reused in other works \cite{kim2021dual}. For each category (Dress, Shirt, Toptee), there is a set of evaluation queries (of size 2017, 2038, and 1961 respectively) as well as a gallery of candidates (of size 3817, 6346, 5373 respectively) that includes a correct match. We report Recall@5 and Recall@10 on each of the three categories.

We note that there exist other evaluation protocols for Fashion IQ \cite{chen2020image, lee2021cosmo}. Some works report numbers for the VAL protocol \cite{chen2020image} instead of (or in addition to) the Original protocol, and so we provide performance according to this protocol as well below. We encourage future works to use the Original protocol for consistent comparisons. \\

\begin{small}
\begin{tabularx}{0.9\columnwidth}{YYYYYYY}
\toprule
\multicolumn{2}{c}{\textbf{Dress}} & \multicolumn{2}{c}{\textbf{Shirt}} & \multicolumn{2}{c}{\textbf{Toptee}} & \multirow{2}{*}{\textbf{Avg}} \\
\textbf{R@10} &   \textbf{@50} &   \textbf{R@10} &   \textbf{@50} &   \textbf{R@10} &   \textbf{@50} &                                   \\
\midrule
37.18 & 64.25 & 34.94 & 61.68 & 42.53 & 71.34 & 51.99 \\
\bottomrule
\end{tabularx}

\end{small}

\paragraph{CR/SR.} These tasks are trained and evaluated separately on classification into 48 categories or 121 subcategories in Fashion-Gen \cite{rostamzadeh2018fashion}. We follow \citet{zhuge2021kaleido} and report accuracy and macro-F1 score.

\paragraph{IC/RIC.} For IC, we evaluate BLEU-4, METEOR, ROUGE-L, and CIDEr scores using \texttt{pycocoevalcap}\footnote{\href{https://github.com/LuoweiZhou/coco-caption}{https://github.com/LuoweiZhou/coco-caption}} following \citet{zhuge2021kaleido}. We use the same evaluation metrics for RIC. As Fashion IQ data contains two relative captions for every (reference, target) pair, we predict two relative captions in RIC, and concatenate them with ``\texttt{and}''. We apply the same procedure to the ground-truth relative captions before calculating the evaluation metrics.

\end{document}